\documentclass[10pt,twocolumn,letterpaper]{article}

\usepackage{cvpr}
\usepackage{times}
\usepackage{epsfig}
\usepackage{graphicx}
\usepackage{amsmath}
\usepackage{amssymb}
\usepackage{color}
\usepackage{breqn}
\usepackage[ruled,vlined]{algorithm2e}

\DeclareMathOperator{\Ob}{\mathbf{O}}

\DeclareMathOperator{\cbb}{\mathbf{c}}
\DeclareMathOperator{\cpb}{\mathbf{c}^\prime}
\DeclareMathOperator{\ib}{\mathbf{i}}

\DeclareMathOperator{\ub}{\mathbf{u}}
\DeclareMathOperator{\wb}{\mathbf{w}}
\DeclareMathOperator{\thetab}{\mathbf{\Theta}}

\DeclareMathOperator*{\argmax}{\mathop{\mathrm{argmax}}}

\newcommand{\cbr}[1]{\left\{#1\right\}}

\def\pairci{\langle\mbox{\it image}, \mbox{\it\{caption(s)\} }\rangle}

\def\mcic{\textsc{MC}$_{\mbox{\scriptsize IC}}\;$}
\def\mciccoco{\textsc{MCIC-COCO}$\;$}
\def\ffnnb{FFNN$_{\mbox{\scriptsize 2-class}}^{\mbox{\scriptsize argmax 1..5}}\;$}
\def\seqff{Vec2seq+FFNN$\;$}

\cvprfinalcopy 

\def\cvprPaperID{3121} 

\begin{document}

\title{Understanding Image and Text Simultaneously: a Dual Vision-Language Machine Comprehension Task}

\author{Nan Ding \\
Google\\
{\tt\small dingnan@google.com}
\and Sebastian Goodman \\
Google\\
{\tt\small seabass@google.com}\\
\and Fei Sha \\
Google\\
{\tt\small fsha@google.com}\\
\and Radu Soricut \\
Google\\
{\tt\small rsoricut@google.com}
}

\maketitle

\begin{abstract}
We introduce a new multi-modal task for computer systems, posed
as a combined vision-language comprehension challenge:
identifying the most suitable \emph{text} describing a scene, given several
similar options.
Accomplishing the task entails demonstrating comprehension beyond just
recognizing ``keywords'' (or key-phrases) and their corresponding visual
concepts.
Instead, it requires an alignment between the representations
of the two modalities that achieves a visually-grounded
``understanding'' of various linguistic elements and their dependencies.
This new task also admits an easy-to-compute and well-studied metric:
the accuracy in detecting the true target among the decoys.

The paper makes several contributions:
an effective and extensible mechanism for generating decoys from (human-created) image captions;
an instance of applying this mechanism, yielding a large-scale machine
comprehension dataset (based on the COCO images and captions)
that we make publicly available;
human evaluation results on this dataset, informing a performance upper-bound;
and several baseline and competitive learning approaches that illustrate the
utility of the proposed task and dataset in advancing both image and language
comprehension.
We also show that, in a  multi-task learning setting, the performance on the proposed task
is positively correlated with the end-to-end task of image captioning.
\end{abstract}

\section{Introduction}

There has been a great deal of interest in multi-modal artificial intelligence research recently,
bringing together the fields of Computer Vision and Natural Language Processing.
This interest has been fueled in part by the availability of many large-scale image datasets with textual annotations.
Several vision+language tasks have been proposed around these datasets~\cite{hodosh13framing,karpathy2014deep,coco,antol15vqa}.
Image Captioning~\cite{hodosh13framing,donahue2014long,karpathy2014deep,fang2014captions,kiros2014unifying,vinyals2014show,mao15mrnn,xu2015show} and Visual Question Answering~\cite{malinoski14qa,malinowski15neural,tu14video,antol15vqa,yu15madlib,wu16external,ren15visual,gao15machine,yang16san,zhu16visual7w,lin16leverage} have in particular attracted a lot of attention.
The performances on these tasks have been steadily improving, owing much to
the wide use of deep learning architectures~\cite{bengio2009book}.

A central theme underlying these efforts is the use of natural language to identify
how much visual information is perceived and understood by a computer system.
Presumably, a system that understands a visual scene well enough
ought to be able to describe what the scene is about (thus ``captioning'') or
provide correct and visually-grounded answers when queried (thus ``question-answering'').

In this paper, we argue for directly measuring how well
the semantic representations of the visual and linguistic
modalities align (in some abstract semantic space).
For instance, given an image and two captions --
a correct one and an incorrect  yet-cunningly-similar one --
can we both qualitatively and quantitatively measure the extent to which
humans can dismiss the incorrect one but computer systems blunder?
Arguably, the degree of the modal alignment is a strong indicator of task-specific
performance on any vision+language task.
Consequentially, computer systems that can learn to maximize and exploit such alignment
should outperform those that do not.

We take a two-pronged approach for addressing this issue.
First, we introduce a new and challenging Dual Machine Comprehension (DMC) task,
in which a computer system must identify the most suitable textual description
from several options: one being the target and the others being ``adversarialy''-chosen decoys.
All options are free-form, coherent, and fluent sentences
with \emph{high degrees of semantic similarity} (hence, they are ``cunningly similar'').
A successful computer system has to demonstrate comprehension beyond just
recognizing ``keywords'' (or key phrases) and their corresponding visual concepts; they
must arrive at a coinciding and visually-grounded understanding of
various linguistic elements and their dependencies.
What makes the DMC task even more appealing is that
it admits an easy-to-compute and well-studied performance metric:
the accuracy in detecting the true target among the decoys.
Second, we illustrate how solving the DMC task benefits related vision+language tasks.
To this end, we render the DMC task as a classification problem, and
incorporate it in a multi-task learning framework for end-to-end training of joint objectives.

Our work makes the following contributions:
(1) an effective and extensible algorithm for generating decoys from human-created image captions (Section~\ref{sec:creation});
(2) an instantiation of applying this algorithm to the COCO dataset~\cite{coco}, resulting in a large-scale
dual machine-comprehension dataset that we make publicly available (Section~\ref{sec:mcic-coco});
(3) a human evaluation on this dataset, which provides an upper-bound on performance (Section~\ref{sec:human_eval});
(4) a benchmark study of baseline and competitive learning approaches (Section~\ref{sec:results}), which underperform humans by a substantial gap (about 20\% absolute);
and (5) a novel multi-task learning model that simultaneously learns to solve the DMC task and
the Image Captioning task (Sections~\ref{sec:seq+ffnn} and \ref{sec:lambda_gen}).

Our empirical study shows that performance on the DMC task positively correlates with performance on the Image Captioning task.
Therefore, besides acting as a standalone benchmark, the new DMC task can be useful in improving other complex vision+language tasks.
Both suggest the DMC task as a fruitful direction for future research.


\section{Related work}
Image understanding is a long-standing challenge in computer vision.
There has recently been a great deal of interest in bringing together vision and language understanding.
Particularly relevant to our work are image captioning (IC) and visual question-answering (VQA).
Both have instigated a large body of publications, a detailed exposition of which is beyond the scope of this paper.
Interested readers should refer to two recent surveys~\cite{bernardi16survey,wu16survey}.

In IC tasks, systems attempt to generate a fluent and correct sentence describing an input image.
IC systems are usually evaluated on how well the generated descriptions align with human-created captions (ground-truth).
The language generation model of an IC system plays a crucial role; it is often trained such that the probabilities of the ground-truth captions are maximized (MLE training),
though more advanced methods based on techniques borrowed from Reinforcement Learning have been proposed~\cite{mixer15}.
To provide visual grounding, image features are extracted and injected into the language model.
Note that language generation models need to both decipher the information encoded in the visual features,
and model natural language generation.

In VQA tasks, the aim is to answer an input question correctly with respect to a given input image.
In many variations of this task, answers are limited to single words or a binary response (``yes'' or ``no'')~\cite{antol15vqa}.
The Visual7W dataset~\cite{zhu16visual7w} contains anaswers in a richer format such as phrases, but limits questions to ``wh-''style (what, where, who, etc).
The Visual Genome dataset~\cite{krishnavisualgenome}, on the other hand, can potentially define more complex questions and answers due to its extensive textual annotations.

Our DMC task is related but significantly different.
In our task, systems attempt to discriminate the best caption for an input image from a set of captions --- all but one are decoys.
Arguably, it is a form of VQA task, where the same default (thus uninformative) question is asked:
\emph{Which of the following sentences best describes this image?}
However, unlike current VQA tasks,
choosing the correct answer in our task entails a deeper ``understanding'' of the available answers.
Thus, to perform well, a computer system needs
to understand both complex scenes (visual understanding) and complex sentences (language understanding),
\emph{and} be able to reconcile them.

The DMC task admits a simple classification-based evaluation metric:
the accuracy of selecting the true target.
This is a clear advantage over the IC tasks, which often rely on imperfect metrics
such as BLEU~\cite{papineni-etal:2002}, ROUGE~\cite{lin-och:2004}, METEOR~\cite{meteor}, CIDEr~\cite{cider}, or SPICE~\cite{spice}.

Related to our proposal is the work in~\cite{hodosh13framing}, which frames image captioning as a ranking problem. While both share the idea of selecting captions from a large set,
our framework has some important and distinctive components.
First, we devise an algorithm for smart selection of candidate decoys, with the goal of selecting those that are
sufficiently similar to the true targets to be challenging,
and yet still be reliably identifiable by human raters.
Second, we have conducted a thorough human evaluation in order to establish a performance ceiling,
while also quantifying the level to which current learning systems underperform.
Lastly, we show that there exists a positive correlation between the performance on the DMC task and the performance on related vision+language tasks
by proposing and experimenting with a multi-task learning model.  Our work is also substantially different from their more recent work~\cite{hodosh16eval}, where only one decoy is considered and its generation is either random, or focusing on visual concept similarity (``switching people or scenes'') instead of our focus on both linguistic surface and paragraph vector embedding similarity.


\section{The Dual Machine Comprehension Task}

\subsection{Design overview}

We propose a new multi-modal machine comprehension task  to examine how well visual and textual semantic understanding are aligned.  Given an image, human evaluators or machines must accurately identify the best sentence describing the scene from several decoy sentences. Accuracy on this task is defined as the percentage that the true targets are identified.

It seems straightforward to construct a dataset for this task, as there are several existing datasets which are composed of images and their (multiple) ground-truth captions, including the popular COCO dataset~\cite{coco}.  Thus, for any given image, it appears that one just needs to use the captions corresponding to other images as decoys. However, this na\"{i}ve approach could be overly simplistic as it is provides no control over the properties of the decoys.

Specifically, our desideratum is to recruit \emph{challenging} decoys that are sufficiently similar to the targets. However, for a small number of decoys, e.g. 4-5, randomly selected captions could be significantly different from the target. The resulting dataset would be too ``easy'' to shed any insight on the task.  Since we are also interested in human performance on this task, it is thus impractical to increase the number of decoys to raise the difficulty level of the task at the expense of demanding humans to examine tediously and unreliably  a large number of decoys.  In short, we need an \emph{automatic procedure to reliably create difficult sets of decoy captions} that are sufficiently similar to the targets.

We describe such a procedure in the following. While it focuses on identifying decoy captions, the main idea is potentially adaptable to other settings. The algorithm is flexible in that the ``difficulty" of the dataset can be controlled to some extent through the algorithm's parameters.

\subsection{Algorithm to create an MC-IC dataset}
\label{sec:creation}

The main idea behind our algorithm is to carefully define a ``good decoy''.  The algorithm exploits recent advances in paragraph vector (PV) models~\cite{le-mikolov:2014}, while also using linguistic surface analysis to define similarity between two sentences. Due to space limits, we omit a detailed introduction of the PV model. It suffices to note that the model outputs a continuously-valued embedding for a sentence, a paragraph, or even a document.

The pseudo-code is given in Algorithm~\ref{aMCIC} (the name \textsc{MC-IC} stands for ``Machine-Comprehension for Image \& Captions''). As input, the algorithm takes a set $C$ of $\pairci$ pairs\footnote{On the order of at least hundreds of thousands of examples; smaller sets result in less challenging datasets.}, as those extracted from a variety of publicly-available corpora, including the COCO dataset~\cite{coco}. The output of the algorithm is the \mcic set.


\begin{algorithm}[t]
\caption{\textsc{MC-IC}($C$, $N$, $\mbox{\it Score}$)}
\label{aMCIC}
\SetAlgoLined
\KwResult{Dataset \mcic}
$\textsf{PV} \gets \textsc{Optimize-PV}(C)$ \\
$\lambda \gets \textsc{Optimize-Score}(\textsf{PV}, C, \mbox{\it Score})$ \\
\mcic$\gets \emptyset$ \\
$nr\_decoys = 4$ \\
\For{$\langle \ib_i, \cbb_i \rangle \in C$}{
  $A\gets []$ \\
  $T_{\cbb_i} \gets \textsf{PV}(\cbb_i)[1..N]$ \\
  \For{$\cbb_d\in T_{\cbb_i}$}{
    $score \gets \mbox{\it Score}(\textsf{PV}, \lambda, \cbb_d, \cbb_i)$ \\
    \If{$score > 0$}{
        $A.\mbox{\bf append}(\langle score, \cbb_d\rangle)$
      }
    }
  \If{$|A| \geq nr\_decoys$}{
      $R\gets \mbox{\bf descending-sort}(A)$ \\
      \For{$l \in [1..nr\_decoys]$}{
        $\langle score, \cbb_d\rangle\gets R[l]$ \\
        \mcic$\gets$ \mcic$\cup\{(\langle \ib_i, \cbb_d\rangle, \mbox{\bf false})\}$ \\
      }
      \mcic$\gets$ \mcic$\cup\{(\langle \ib_i, \cbb_i \rangle, \mbox{\bf true})\}$ \\
    }
}
\end{algorithm}

Concretely, the \textsc{MC-IC} Algorithm has three main arguments:
a dataset $C = \{ \langle \ib_i, \cbb_i \rangle | 1 \leq i \leq m\}$ where $\ib_i$ is an image and $\cbb_i$ is its ground-truth caption\footnote{For an image with multiple ground-truth captions, we split it to multiple instances with the same image for each one of the ground-truth captions; the train/dev/test splits are done such that they contain disjoint {\em image} sets, as opposed to disjoint {\em instance} sets.}; an integer $N$ which controls the size of $\cbb_i$'s neighborhood in the embedding space defined by the paragraph vector model \textsf{PV}; and a function \textsf{Score} which is used to score the $N$ items in  each such neighborhood.

The first two steps of the algorithm tune several hyperparameters. The first step finds optimal settings for the
\textsf{PV} model given the dataset $C$. The second  finds a weight parameter
$\lambda$ given \textsf{PV}, dataset $C$, and the \textsf{Score}  function.
These hyperparameters are dataset-specific. Details are discussed in the next section.

The main body of the algorithm, the outer $\textbf{for}$ loop, generates a set of $nr\_decoys$ (4 here) decoys for each ground-truth caption. It accomplishes this by first extracting $N$ candidates from the \textsf{PV} neighborhood of the ground-truth caption, excluding those that belong to the same image. In the inner $\textbf{for}$ loop, it computes the similarity of each candidate to the ground-truth and stores them in a list $A$.  If enough candidates are generated, the list is sorted in descending order of score. The top $nr\_decoys$ captions are marked as ``decoys'' (\ie \textbf{false}), while the ground-truth caption is marked as ``target'' (\ie \textbf{true}).

The score function $\textsf{Score}(\textsf{PV}, \lambda, \cpb, \cbb)$ is a crucial component of the decoy selection mechanism. Its definition leverages our linguistic intuition by combining linguistic surface similarity,  $\textsf{sim}_{\textsc{surf}}(\cpb, \cbb)$, with the similarity suggested by the embedding model, $\mbox{\textsf{sim}}_{\textsf{PV}}(\cpb, \cbb)$:
\begin{equation}
\textsf{Score}\!=\!
\left\{\!\!
\arraycolsep=1.5pt
\begin{array}{ll}
0 & \! \mbox{if \textsf{sim}}_{\textsc{surf}}\!\ge\!L \\
\lambda\, \mbox{\textsf{sim}}_{\textsf{PV}}\!+\!(1\!-\!\lambda)\, \mbox{\textsf{sim}}_{\textsc{surf}}  & \text{otherwise}
\end{array}
\right.
\label{eq:score}
\end{equation}
where the common argument $(\cpb, \cbb)$ is omitted. The higher the similarity score, the more likely that $\cpb$ is a good decoy for $\cbb$. Note that if the surface similarity is above the threshold $L$, the function returns 0, flagging that the two captions are too similar to be used as a pair of target and decoy.

In this work, $\mbox{\textsf{sim}}_{\textsc{surf}}$ is computed as the BLEU score between the inputs~\cite{papineni-etal:2002} (with the brevity penalty set to 1). The embedding similarity, $\mbox{\textsf{sim}}_{\textsf{PV}}$, is computed as the cosine similarity between the two in the PV embedding space.

\begin{figure*}[th]
\centering
\begin{minipage}[t]{0.45\textwidth}
\vspace{0pt}
\centering
\includegraphics[width=0.4\textwidth]{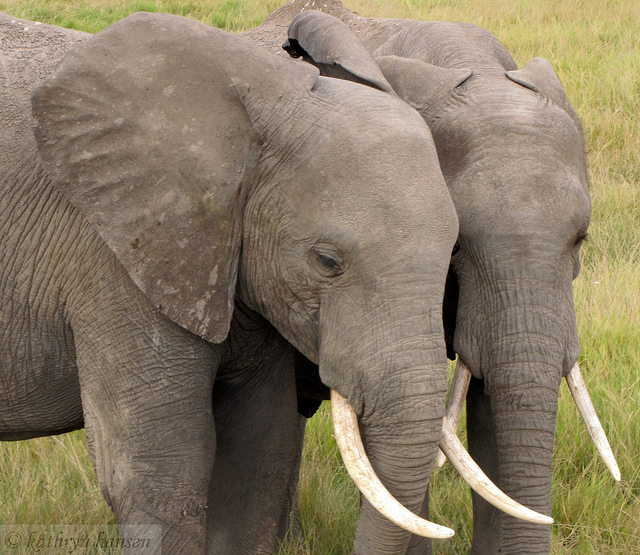}
\end{minipage}%
\begin{minipage}[t]{0.45\textwidth}
\vspace{1em}
\begin{tabular}{l}
1. a herd of giraffe standing next to each other in a dirt field \\
2. a pack of elephants standing next to each other \\
3. animals are gathering next to each other in a dirt field \\
4. three giraffe standing next to each other on a grass field \\
5. {\bf two elephants standing next to each other in a grass field}\\
\end{tabular}
\end{minipage}%
\\
\begin{minipage}[t]{0.90\textwidth}
\vspace{1em}
\begin{tabular}[b]{l}
{\bf Instructions:} \\
Captions 1 and 4 are clearly incorrect. They do not match up with the image at all. \\
Caption 2 calls the elephants a "pack", which is vague and a bit subjective. It does not mention the grass at all. \\
Caption 3 only uses the word "animals" to describe what is in the picture, when the picture clearly shows elephants. \\
Caption 5 gives the exact count and correct animal type and even mentions the grass field. It is a more accurate, \\
descriptive, and objective caption than the other options. \\
\end{tabular}
\end{minipage}
\\[0.5em]
\begin{minipage}[t]{0.45\textwidth}
  \vspace{0pt}
  \centering
\includegraphics[width=0.4\textwidth]{coco-food.jpg}
\end{minipage}%
\begin{minipage}[t]{0.45\textwidth}
  \vspace{0.5em}
\begin{tabular}{l}
1. a meal covered with a lot of broccoli and tomatoes \\
2. a pan filled with a mixture of vegetables and meat \\
3. {\bf a piece of bread covered in a meat and sauces} \\
4. a pizza smothered in cheese and meat with french fries \\
5. a plate of fries and a sandwich cut in half \\
\end{tabular}
\end{minipage}%
\\
\begin{minipage}[t]{0.90\textwidth}
  \vspace{1em}
\centering
\begin{tabular}{l}
{\bf Instructions:} \\
Caption 3 does not mention the fork, the plate, or the eggs, but it is still the best option because the other captions are \\
inaccurate. Captions 1, 2, 4, and 5 all describe items not present in the picture, such as broccoli, a pan, cheese, or fries. \\
\end{tabular}
\end{minipage}
\\[0.25em]
\begin{minipage}[t]{0.45\textwidth}
  \vspace{0pt}
\centering
\includegraphics[width=0.4\textwidth]{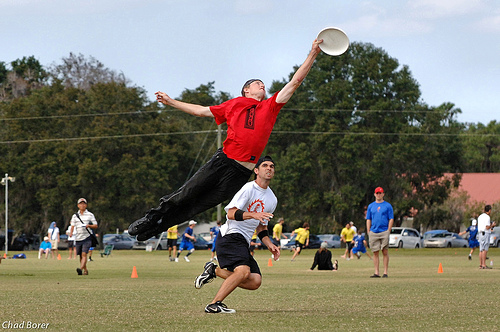}
\end{minipage}%
\begin{minipage}[t]{0.45\textwidth}
  \vspace{0pt}
\begin{tabular}{l}
1. the man in the picture is reaching toward a frisbee \\
2. a middle aged man in a field tossing a frisbee \\
3. a woman in stance to throw a frisbee \\
4. {\bf a man dives for a catch in this ultimate frisbee match } \\
5. there is a male tennis player playing in a match
\end{tabular}
\end{minipage}%
\\
\begin{minipage}[t]{0.90\textwidth}
  \vspace{0pt}
\centering
\begin{tabular}{l}
{\bf Instructions:} \\
Caption 5 is clearly incorrect (the game being played here is not tennis).
Captions 2 and 3 describe the act of tossing, \\
but the picture shows the act of catching, so these captions are both inaccurate.
Captions 1 and 4 seem close, but the \\
phrase "dives for a catch" is more descriptive than the phrase "reaching toward a frisbee". In addition, Caption 4 \\
mentions the name of the game that they are playing, so it better informs the reader about what is happening in the \\
rest of the image than the other caption does.
\end{tabular}
\end{minipage}
\\[0.5em]
\caption{Examples of instances from the \mciccoco dataset (the ground-truth
is in {\bf bold} face), together with rater instructions.}
\label{fig:rater-training}
\end{figure*}

\begin{table}[t]
  \small
  \begin{center}
    \begin{tabular}{r | r r r r }
      Split & dev & test & train & total  \\ \hline
      \#unique\_images   & 2,000  & 2,000 & 110,800 & 114,800\\
      \# instances & 9,999 & 10,253 &  554,063 &  574,315
    \end{tabular}
  \end{center}
  \caption{\mciccoco dataset descriptive statistics}
  \label{table:mcic_splits}
\end{table}

\subsection{The \mciccoco dataset}
\label{sec:mcic-coco}

We applied the \textsc{MC-IC} Algorithm to the COCO dataset~\cite{coco} to generate
a dataset for the visual-language dual machine comprehension task. The dataset is called \mciccoco and it is made
publicly available\footnote{\tt http://www.github.com/google/mcic-coco}.  We describe the details of this dataset below.

We set the neighborhood size at $N=500$, and the threshold at $L=0.5$ (see Eq.~\ref{eq:score}). As the COCO dataset has a large body of images (thus captions) focusing on a few categories (such as sports activities), this threshold is important in discarding significantly similar captions to be decoys -- otherwise, even human annotators will experience difficulty in selecting the ground-truth captions.

The hyperparameters of the \textsf{PV} model, {\tt dim} (embedding dimension) and
{\tt epochs} (number of training epochs), are optimized in the $\textsc{Optimize-PV}$ step of
the \textsc{MC-IC} Algorithm. The main idea is to learn embeddings such that
ground-truth captions from the same image have similar embeddings.
Concretely, the optimization step is a grid-search over
the hyper-parameters of the PV-DBOW model~\cite{le-mikolov:2014},
which we train using a softmax loss. Since there are multiple ground-truth captions associated with each image,
the dataset is denoted by $C = \{ \langle \ib_{r_c}, \cbb_{r_c} \rangle | 1 \leq r \leq n, 1 \leq c \leq s_r \}$, where $r$ is the index for each unique image ($\ib_{r_c} \equiv \ib_r$), $n$ is the total number images and $s_r > 1$ is the number
of unique captions for image $r$. The total number of data examples $m = \sum_{r=1}^n s_r$. Here
the hyper-parameters are searched on a grid  to minimize
``multiple ground-truth score'' rank (mgs-rank):
the average rank (under the cosine-distance score)
between $\cbb_{r_c}$ and $\{ \cbb_{r_l} | 1 \leq l \leq s_r, l\not= c\}$.
The lower the mgs-rank, the better the resulting paragraph vector model is
at modeling multiple ground-truths for a given image as being similar.
As such, our grid-search over the \mciccoco dev dataset yields
a minimum mgs-rank at {\tt dim}=1024 and {\tt epochs}=5.

Similarly, the
$\textsc{Optimize-Score}(\textsf{PV}, \textsf{Score})$ step
is a grid-search over the $\lambda$ parameter of the \textsf{Score}
function, given a paragraph vector embedding model \textsf{PV} and a dataset $C$ of captions and images, as before. A well-chosen $\lambda$ will ensure  the multiple ground-truth captions for the same image will be measured with high degree of similarity with the \textsf{Score} function.
The $\lambda\in [0,1]$ parameter is searched on a grid to minimize
the ``weighted multiple ground-truths score'' rank (wmgs-rank):
the average rank (under the \textsf{Score})
between $\cbb_{r_c}$ and $\{ \cbb_{r_l} | 1 \leq l \leq s_r, l\not= c\}$,
relative to the top $N$-best closest-cosine neighbors in \textsf{PV}.
For example, if given five ground-truths for image $\ib_r$,
and when considering $\cbb_{r_1}$,
ground-truths $\cbb_{r_2}$ to $\cbb_{r_5}$
are ranking at \#4, \#10, \#16, and \#22
(in top-500 closest-cosine neighbors in \textsf{PV}),
then wmgs-rank$(\cbb_{r_1})=13$ (the average of these ranks).
Our grid-search over the \mciccoco dev dataset yields
a minimum wmgs-rank at $\lambda$=0.3.

The resulting \mciccoco dataset has 574,315 instances that are in the format of
$\{i: (\langle \ib_i, \cbb_i^j \rangle , \mbox{label}_i^j), j = 1 \ldots 5\}$ where $\mbox{label}_i^j\in \{\mbox{\bf true}, \mbox{\bf false}\}$.
For each such instance, there is one and only one $j$ such that the label is \textbf{true}.
We have created a train/dev/test split such that all of the instances for the same image occur in the same split. Table~\ref{table:mcic_splits} reports the basic statistics for the dataset.

\begin{table}[t]
  \small
  \begin{center}
    \begin{tabular}{r | r | c }
      Correct responses    & \# instances & Accuracy\% \\ \hline
      3 out of 3           & 673          & 67.3 \\
      at least 2 out of 3  & 828          & \textbf{82.8} \\
      at least 1 out of 3  & 931          & 93.1 \\
       0 out of 3          &  69          &  0.0  \\
    \end{tabular}
  \end{center}
  \caption{Human performance on the DMC task with the \mciccoco dataset. \textbf{Bold} denotes the performance ceiling.}
  \label{table:human_eval_instances}
\end{table}

\subsection{Human performance on \mciccoco}
\label{sec:human_eval}

\noindent
\textbf{Setup}\ To measure how well humans can perform on the DMC task, we randomly drew
1,000 instances from the \mciccoco dev set and submitted those instances to human
``raters''\footnote{Raters are vetted, screened and tested before working on any tasks;
requirements include native-language proficiency level.} via a crowd-sourcing platform.

Three independent responses from 3 different rates were gathered for each instance, for a total of 3,000 responses. To ensure diversity, raters were prohibited from evaluating more than six instances or from responding to the same task instance twice. In total, 807 distinct raters were employed.

Raters were shown one instance at a time. They were shown the image and the five caption choices
(ground-truth and four decoys, in randomized order) and were instructed to choose the best caption for the image.
Before starting evaluation, the raters were trained with sample instances from the \emph{train} dataset, disjoint from the \emph{dev} dataset on which their performance data were collected. The training process presents
 an image and five sentences, of which the ground-truth caption is highlighted.
In addition, specific instructions and clarification were given to the raters on how to choose the best caption for the image.
In Figure~\ref{fig:rater-training}, we present three instances on how the rater instructions were presented for rater training.

\begin{figure*}[th]
\centering
\begin{minipage}[t]{0.45\textwidth}
  \vspace{0pt}
\centering
\includegraphics[width=0.4\textwidth]{coco-beach-horseriding.jpg}
\end{minipage}%
\begin{minipage}[t]{0.45\textwidth}
  \vspace{0pt}
\begin{tabular}{l}
1. three bikes on the shore while people talk on a small boat \\
2. three people and a dog are running on the beach \\
3. {\bf three people ride horses along the beach while a dog follows} \\
4. two people on horseback ride along a beach \\
5. two people on horses trot along a sandy beach \\
\end{tabular}
\end{minipage}%
\\[0.25em]
\begin{minipage}[t]{0.45\textwidth}
  \vspace{0pt}
\centering
\includegraphics[width=0.4\textwidth]{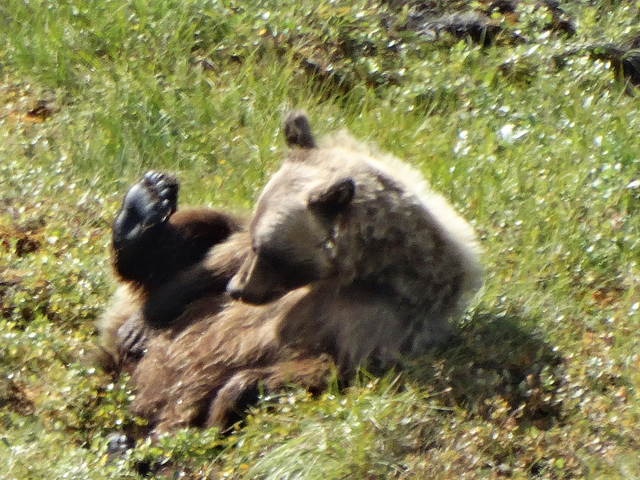}
\end{minipage}%
\begin{minipage}[t]{0.45\textwidth}
\vspace{0.25em}
\begin{tabular}{l}
1. a brown bear is standing in the grass \\
2. a brown bear standing in the water of a river \\
3. a dark brown bear standing in the woods \\
4. {\bf a small brown bear rolling in the grass} \\
5. a small brown bear standing in the dirt \\
\end{tabular}
\end{minipage}%
\\[0.25em]
\begin{minipage}[t]{0.45\textwidth}
\vspace{0pt}
  \centering
\includegraphics[width=0.4\textwidth]{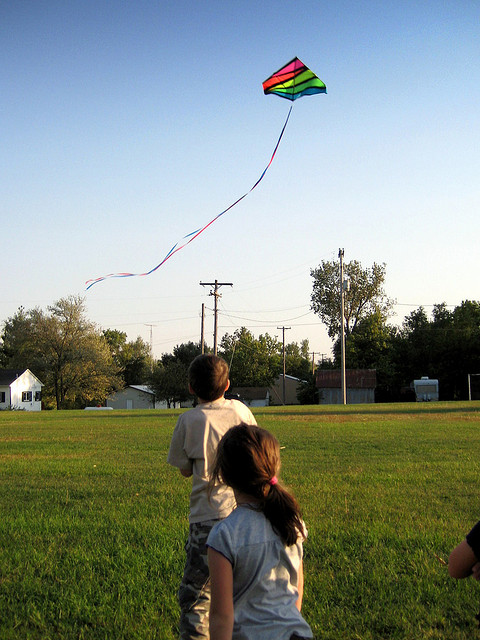}
\end{minipage}%
\begin{minipage}[t]{0.45\textwidth}
  \vspace{2em}
\begin{tabular}{l}
1. a family is playing the wii in a house \\
2. a man playing with a frisbee in a park \\
3. a small boy playing with kites in a field \\
4. three women play with frisbees in a shady park\\
5. {\bf boy flies a kite with family in the park} \\
\end{tabular}
\end{minipage}%
\vspace{1em}
\caption{Examples of instances from the \mciccoco dataset (the ground-truth
is in {\bf bold} face).
The correct answers for the first two examples are relatively obvious to humans,
but less so to computer systems.
The third example illustrates one of the difficult cases in which
the humans annotators did not agree (option 3. was also chosen).}
\label{fig:examples}
\end{figure*}

\noindent
\textbf{Quantitative results}\ We assessed human performance in two metrics: (1)
Percentage of correct rater responses (1-human system): \textbf{81.1\%} (2432  out of 3000);
(2) Percentage of instances with at least $50\%$ (\ie 2)  correct responses (3-human system): \textbf{82.8\%} (828 out of 1000).

Table~\ref{table:human_eval_instances} gives a detailed breakdown
on the statistics related to the inter-rater (dis)agreement.
The first row, with accuracy at 67.3\%, suggests that this is the level at which the
correct answer is obvious (\ie, percentage of ``easy'' instances).
The second row, at 82.8\%, indicates that this is the performance ceiling in terms of accuracy
that can be expected for the \mciccoco dataset; at the same time, it suggests that the
difference between 67.3\% and 82.8\% (i.e., about 15\% of instances) is caused by ``difficult'' instances.
Finally, the third row, at 93.1\%, indicates that the level of ``unanswerable'' instances
is somewhere in the 10\%-15\% range (combining the increase from 82.8\% to 93.1\% and the remaining 6.9\% that no one gets right).

We will investigate those instances in detail in the future.  The COCO dataset has a significant number of captions
that fit more than one image in the dataset, given the biased concentration on certain categories. Thus, we suspect that even with our threshold-check (cf. the introduction of $L$ in Eq.~\ref{eq:score}), our procedure might have failed to filter out some impossible-to-distinguish decoys.

\noindent
\textbf{Qualitative examples}\ We present in Figure~\ref{fig:examples} several example instances
from the \mciccoco dataset.
The first example illustrates how certain aspects of VQA are subsumed by the DMC task:
in order to correctly choose answer 3, a system needs to implicitly answer questions like
``how many people are in the image?'' (answer: three, thus choices 4. and 5. are wrong), and
``what are the people doing?'' (answer: riding horses, thus choices 1. and 2. are wrong).
The second example illustrates the extent to which a successful computer system needs to be
able to differentiate between ``standing'' and ``rolling'' in a visually-grounded way, presumably
via a pose model~\cite{yao-li:2010} combined with a translation model between poses and their verbal correspondents.
Last but not least, the third examples illustrates a difficult case, which led to human annotator
disagreement in our annotation process (both choice 3. and 5. were selected by different annotators).


\section{Learning Methods}
\label{sLearning}

We describe several learning methods for the dual machine comprehension (DMC) task with the \mcic dataset.  We start with linear models which will be used as baselines. We then present several neural-network based models. In particular, we describe a novel, hybrid neural network model that combines the feedforward architecture and the seq2seq architecture~\cite{sutskever-etal:2014} for multi-task learning of the DMC task and the image captioning task. This new model achieves the best performance in both tasks.

\subsection{Linear models as baselines}

\noindent
\textbf{Regression}\ To examine how well the two embeddings are aligned in ``semantic understanding space'', a simple approach is to assume that the learners do not have access to the decoys.
Instead, by  accessing the ground-truth captions only, the models
learn a linear regressor from the image embeddings to the target captions' embeddings
(``forward regression''), or from the captions to the images
(``backward regression'').
With the former approach, referred as \textsf{Baseline-I2C}, we check whether the predicted
caption for any given image is closest to its true caption.
With the latter, referred as \textsf{Baseline-C2I}, we check whether the predicted image embedding by the ground-truth caption is the closest among predicted ones by decoy captions to the real image embeddings.

\noindent
\textbf{Linear classifier}\ Our next approach \textsf{Baseline-LinM} is a linear  classifier learned to
discriminate true targets from the decoys. Specifically, we learn a linear discriminant function $f(\ib, \cbb; \thetab) = \ib^{\top}\thetab \cbb$ where $\thetab$ is a matrix measuring
the compatibility between two types of  embeddings, cf. ~\cite{frome2013devise}.
The loss function is then given by
\begin{equation}
  L(\thetab) = \sum_i [ \max_{j \ne j^*} f(\ib_i, \cbb_i^j; \thetab) - f(\ib_i, \cbb_i^{j^*}; \thetab)  ]_+
\end{equation}
where $[\ ]_{+}$ is the hinge function and $j$ indexes over all the
available decoys and $i$ indexes over all training instances.
The optimization tries to increase the gap between the target $\cbb_i^{j^*}$ and the
worst ``offending'' decoy.
We use stochastic (sub)gradient methods to optimize $\thetab$,
and select the best model in terms of accuracy on the \mciccoco development set.


\subsection{Feedforward Neural Network (FFNN) models}
\label{sec:ffnn}
To present our neural-network--based models,
we use the following notations. Each training instance pair is a tuple $\langle \ib_i,  \cbb_i^j\rangle$,
where $\ib$ denotes the image, and $\cbb_i^j$ denotes the caption options, which can either be the target or the decoys. We use a binary variable $y_{ijk} \in \{0,1\}$ to denote whether $j$-th caption of the
instance $i$ is labeled as $k$, and $\sum_k y_{ijk} = 1$.

We first employ the standard feedforward neural-network models
to solve the DMC task on the \mciccoco dataset.
For each instance pair $\langle \ib_i, \cbb_i^j\rangle$, the input to the neural network is an embedding
tuple $\langle \text{DNN}(\ib_i; \Gamma), \text{Emb}(\cbb_i^j; \Omega) \rangle$,
where $\Gamma$ denotes the parameters of a deep convolutional neural network $\text{DNN}$. $\text{DNN}$ takes an image and outputs an image embedding vector. $\Omega$ is the embedding matrix, and
$\text{Emb}(.)$ denotes the mapping from a list of word IDs to a list of
embedding vectors using $\Omega$. The loss function for our FFNN is given by:
\begin{dmath}
L(\Gamma, \Omega,\ub)\!=\! \sum_{i, j, k} y_{ijk} \log \; \text{FN}_k(\text{DNN}(\ib_i; \Gamma), \text{Emb}(\cbb_i^j; \Omega); \ub)
\label{eq:closs}
\end{dmath}
\noindent
where $\text{FN}_k$ denotes the $k$-th output of a feedforward neural network,
and $\sum_k \text{FN}_k(.) = 1$.
Our architecture uses a two hidden-layer fully connected network
with Rectified Linear hidden units, and a softmax layer on top.

The formula in Eq.~\ref{eq:closs} is generic with respect to the number of classes.
In particular, we consider a 2-class--classifier ($k\in\{0, 1\}$, 1 for 'yes', this is a correct answer; 0 for 'no', this is an incorrect answer),
applied independently on all the $\langle \ib_i, \cbb_i^j\rangle$ pairs and apply one FFNN-based binary classifier for each;
the final prediction is the caption with the highest 'yes' probability among all instance pairs belonging to instance $i$.

\subsection{Vec2seq + FFNN Model}
\label{sec:seq+ffnn}
\begin{figure}[th]
\centering
\includegraphics[width=3.5in]{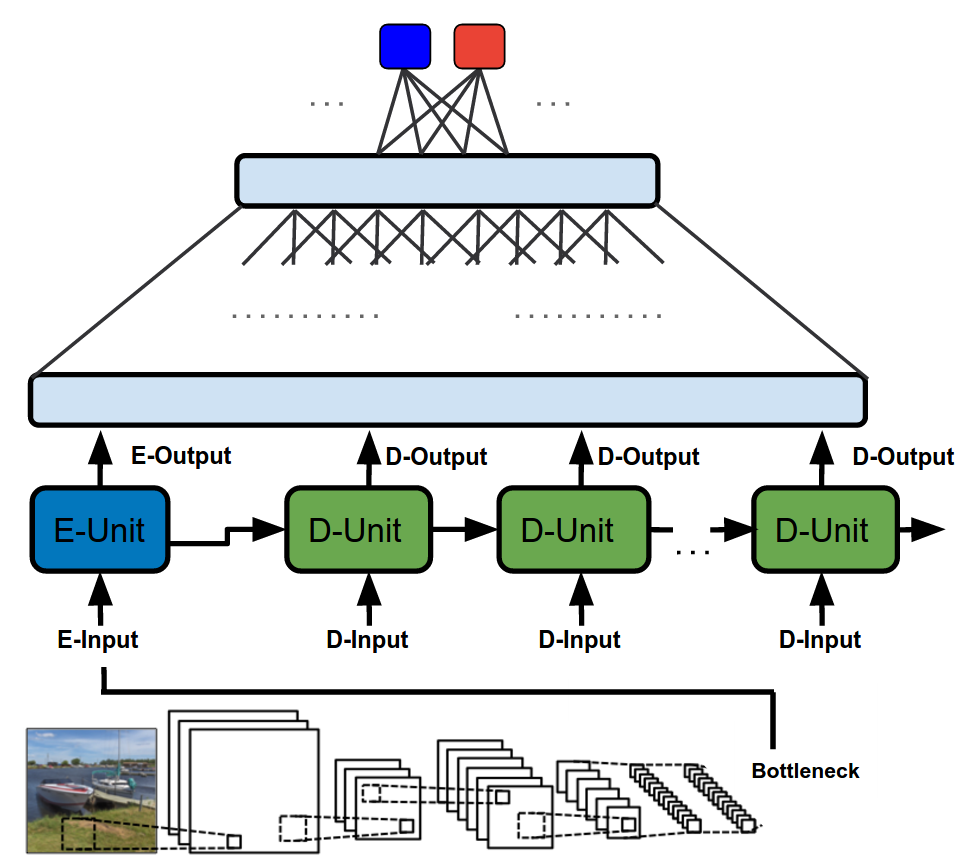}
\caption{Vec2seq + FFNN model architecture.}
\label{agmc_diagram}
\end{figure}

We describe here a hybrid neural-network model that combines a recurrent neural-network
with a feedforward one. We encode the image into a single-cell RNN encoder, and the caption into an RNN decoder. Because the first sequence only contains one cell, we call this model a vector-to-sequence (Vec2seq) model as a special case of Seq2seq model as in ~\cite{sutskever-etal:2014,bahdanau-etal:2015}.
The output of each unit cell of a Vec2seq model (both on the encoding side and the decoding side)
can be fed into an FFNN architecture for binary classification.
See Figure~\ref{agmc_diagram} for an illustration of the Vec2seq + FFNN model architecture.

\paragraph{Multi-task learning} In addition to the classification loss (Eq.~\ref{eq:closs}), we also include a loss for generating an output sequence
$\cbb_i^j$ based on an input $\ib_i$ image.
We define a binary variable $z_{ijlv} \in \{0,1\}$ to indicate whether
the $l$th word of $\cbb_i^j$ is equal to word $v$.
$\Ob^d_{ijl}$ denotes the $l$-th output of the decoder of instance pair $\langle \ib_i, \cbb_i^j\rangle$,
$\Ob^e_{ij}$ denotes the output of the encoder, and $\Ob^d_{ij:}$ denotes the concatenation of decoder outputs.

With these definitions, the loss function for the Vec2seq + FFNN model is:
\begin{dmath}
L(\thetab, \wb, \ub) =
\sum_{i, j, k} y_{ijk} \log \; \text{FN}_k(\Ob^e_{ij}(\ib_i, \cbb_i^j; \thetab), \Ob^d_{ij:}(\ib_i, \cbb_i^j; \thetab); \ub)
+ \lambda_{gen} \sum_{i,j, l, v} y_{ij1} z_{ijlv} \log \;\text{softmax}_v(\Ob^d_{ijl}(\ib_i, \cbb_i^j; \thetab); \wb)
\label{eq:mloss}
\end{dmath}
\noindent
where $\sum_v \text{softmax}_v(.) = 1$;
$\thetab$ are the parameters of the Vec2seq model,
which include the parameters within each unit cell,
as well as the elements in the embedding matrices for images and target sequences;
$\wb$ are the output projection parameters that transform the output space of the decoder
to the vocabulary space.
$\ub$ are the parameters of the FFNN model (Eq.~\ref{eq:closs});
$\lambda_{gen}$ is the weight assigned to the sequence-to-sequence generation loss.
Only the true target candidates (the ones with $y_{ij1} = 1$) are included in this loss,
as we do not want the decoy target options to affect this computation.

The Vec2seq model we use here is an instantiation of the attention-enhanced models
proposed in~\cite{bahdanau-etal:2015,chen-etal:2016}. However, our current model does not support location-wise attention, as in
the Show-Attend-and-Tell~\cite{xu-etal:2016} model.
In this sense, our model is an extension of the Show-and-Tell model with a single attention state
representing the entire image, used as image memory representation for all decoder decisions.
We apply Gated Recurrent Unit (GRU) as the unit cell~\cite{cho-etal:2014}.
We also compare the influence on performance of the $\lambda_{gen}$ parameter.


\section{Experiments}
\label{sec:results}
\subsection{Experimental Setup}

\noindent
\textbf{Baseline models}\ For the baseline models, we use the 2048-dimensional outputs of
Google-Inception-v3~\cite{inception-v3} (pre-trained on ImageNet ILSSVR 2012) to represent the images,
and 1024-dimensional paragraph-vector embeddings (section~\ref{sec:creation})
to represent captions.
To reduce computation time, both are reduced to 256-dimensional vectors
using random projections.

\noindent
\textbf{Neural-nets based models}\ The experiments with these models are done using the Tensorflow package~\cite{tensorflow2015-whitepaper}.
The hyper-parameter choices are decided using the hold-out development
portion of the \mciccoco set.
For modeling the input tokens, we use a vocabulary size of 8,855 types, selected as the most frequent tokens over the captions from the COCO training set
(words occurring at least 5 times).
The models are optimized using ADAGRAD with an initial learning rate of 0.01,
and clipped gradients (maximum norm 4).
We run the training procedures for $3,000,000$ steps, with a mini-batch size of 20. We use 40 workers for computing the updates,
and 10 parameter servers for model storing and (asynchronous and distributed)
updating.

We use the following notations to refer to the neural network models:
\ffnnb refers to the version of feedforward neural network architecture
with a 2-class--classifier ('yes' or 'no' for answer correctness),
over which an $\argmax$ function computes a 5-way decision
(i.e., the choice with the highest 'yes' probability);
we henceforth refer to this model simply as FFNN.

The \seqff refers to the hybrid model described in Section~\ref{sec:seq+ffnn},
combining Vec2seq and  \ffnnb.
The RNN part of the model uses a two-hidden--layer GRU
unit-cell~\cite{cho-etal:2014}
configuration, while the FFNN part uses a two-hidden--layer architecture.
The $\lambda_{gen}$ hyper-parameter from the loss-function
$L(\thetab, \wb, \ub)$ (Eq.~\ref{eq:mloss}) is by default set to 1.0
(except for Section~\ref{sec:lambda_gen} where we directly measure its
effect on performance).

\noindent
\textbf{Evaluation metrics}\ The metrics we use to measure performance come in two flavors.
First, the accuracy in detecting (the index of) the true target among the decoys
provides a direct way of measuring the performance level on the comprehension task.
We use this metric as the main indicator of comprehension performance.
Second, because our \seqff models are multi-task models, they can also
generate new captions given the input image.
The performance level for the generation task is measured using the standard
scripts measuring ROUGE-L~\cite{lin-och:2004} and CIDEr~\cite{cider}, using
as reference the available captions from the COCO data
(around 5 for most of the images).
Code for these metrics is available as part of the COCO evaluation
toolkit~\footnote{\tt https://github.com/tylin/coco-caption}.
As usual, both the hypothesis strings and the reference strings are preprocessed:
remove all the non-alphabetic characters; transform all letters to lowercase, and
tokenize using white space; replace all words occurring less than 5 times with an unknown token $\langle\mbox{UNK}\rangle$ (total vocabulary of 8,855 types);
truncate to the first 30 tokens.

\subsection{Results}
\label{sDMCResults}

Table~\ref{table:baselines} summarizes our main results on the comprehension task. We report the accuracies (and their standard deviations) for random choice, baselines, and neural network-based models.

\begin{table}[t]
  \small
  \begin{center}
    \begin{tabular}{ r | c | c c }
      Model                 & dim & Dev & Test\\ \hline
      \textsf{Baseline-I2C}           & 256 & 19.6 $\pm$0.4 & 19.3$\pm$0.4 \\
      \textsf{Baseline-C2I}           & 256 & 32.8 $\pm$0.5 & 32.0$\pm$0.5 \\
      \textsf{Baseline-LinM}         & 256 & 44.6 $\pm$0.5 & 44.5$\pm$0.5 \\ \hline
      FFNN                  & 256 & 56.3 $\pm$0.5 & 55.1$\pm$0.5 \\
      \seqff                & 256 & \textbf{60.5} $\pm$0.5 & \textbf{59.0}$\pm$0.5 \\
    \end{tabular}
  \end{center}
  \caption{Performance on the DMC Task, in accuracies (and standard deviations) on \mciccoco for baselines and NN models.}
  \label{table:baselines}
\end{table}

Interestingly, the \textsf{Baseline-I2C} model performs at the level of
random choice, and much worse than the \textsf{Baseline-C2I model}.
This discrepancy reflects the inherent difficulty in vision-Language tasks:
for each image, there are several possible equally good descriptions,
thus a linear mapping from the image embeddings to the captions might not
be enough -- statistically, the \emph{linear} model will just predict the
mean of those captions.
However, for the reverse direction where the captions are the independent
variables, the learned model does not have to capture the variability
in image embeddings corresponding to the different but equally good captions
-- there is only one such image embedding.

Nonlinear neural networks overcome these modeling
limitations. The results clearly indicate their superiority over the baselines.
The \seqff model obtains the best results, with accuracies of 60.5\% (dev)
and 59.0\% (test); the accuracy numbers indicate that the \seqff architecture
is superior to the non-recursive fully-connected FFNN
architecture (at 55.1\% accuracy on test).
We show next the impact on performance of the embedding dimension and
neural-network sizes, for both the feedforward and the recurrent
architectures.

\subsection{Analysis: embedding dimension and neural-network sizes}
\label{sec:impact_size}
In this section, we compare neural networks models of different sizes.
Specifically, we compare embedding dimensions of
$\cbr{64, 256, 512, 1024, 2048}$,
and two hidden-layer architectures with sizes of
$\cbr{(64, 16), (256, 64), (512, 128), (1024, 256), (2048, 512)}$.

\begin{table}[t]
  \small
  \begin{center}
    \begin{tabular}{r r r | c c}
      dim  & hidden-1 & hidden-2 & Dev & Test\\ \hline
      \multicolumn{5}{c}{FFNN} \\ \hline
        64 &       64 &       16 & \textbf{56.5} & \textbf{53.9} $\pm$0.5 \\
       256 &       64 &       16 & 56.3 & 55.1 $\pm$0.5 \\
       256 &      256 &       64 & 55.8 & 54.3 $\pm$0.5 \\
       512 &      512 &      128 & 54.1 & 52.5 $\pm$0.5 \\
      1024 &     1024 &      256 & 52.2 & 51.3 $\pm$0.5 \\
      2048 &     2048 &      512 & 50.7 & 50.7 $\pm$0.5 \\ \hline
     \multicolumn{5}{c}{\seqff} (with default $\lambda_{gen} = 1.0$) \\ \hline
        64 &       64 &       16 & 55.3 & 54.0 $\pm$0.5 \\
       256 &       64 &       16 & 60.5 & 59.0 $\pm$0.5 \\
       256 &      256 &       64 & 61.2 & 58.8 $\pm$0.5 \\
       512 &      512 &      128 & 61.6 & 59.6 $\pm$0.5 \\
      1024 &     1024 &      256 & 62.5 & 60.8 $\pm$0.5 \\
      2048 &     2048 &      512 & \textbf{63.4} & \textbf{60.8} $\pm$0.5 \\ \hline
    \end{tabular}
  \end{center}
  \caption{The impact of model sizes on \mciccoco accuracy for the FFNN model.}
  \label{table:ffnn_sizes}
\end{table}

The results in Table~\ref{table:ffnn_sizes} illustrate an interesting behavior
for the neural-network architectures.
For the FFNN models, contrary to expectations,
bigger network sizes leads to decreasing accuracy.
On the other hand, for \seqff models, accuracy increases with increased size in
model parameters, up until the embedding dimension of the RNN model matches
the embedding dimension of the Inception model, at 2048.

At accuracy levels of 63.4\% (dev) and 60.8\% (test), this performance
establishes a high-bar for a computer model performance on the DMC task using the \mciccoco dataset.
According to the estimate from Table~\ref{table:human_eval_instances}, this level of performance
is still \emph{significantly} below the 82.8\% accuracy achievable by humans, which makes \mciccoco
a challenging testbed for future models of Vision-Language machine
comprehension.

\subsection{Multi-task learning for DMC and Image Captioning}
\label{sec:lambda_gen}
In this section, we compare models with different values of $\lambda_{gen}$ in Eq.~\ref{eq:mloss}.
This parameter allows for a natural progression from learning for the DMC task only ($\lambda_{gen} = 0$) to focusing on the image captioning loss ($\lambda_{gen} \rightarrow +\infty$). In between the two extremes, we have a multi-task learning objective for jointly learning related tasks.

\begin{table}[htp]
  \small
  \begin{center}
    \begin{tabular}{r | c c | c c | c c}
      $\lambda_{gen}$ & \multicolumn{2}{c}{Acc} & \multicolumn{2}{c}{ROUGE-L} & \multicolumn{2}{c}{CIDEr} \\
            & Dev   & Test        & Dev    & Test   & Dev    & Test \\ \hline
        0.0 & 50.7 & 50.7  $\pm$0.5  & -      & -      & -      & -    \\
        0.1 & 61.1 & 59.0  $\pm$0.5  & 0.517 & 0.511 & 0.901 & 0.865 \\
        1.0 & \textbf{63.4} & \textbf{60.8}  $\pm$0.5  & 0.528 & 0.518 & 0.972 & 0.903 \\
        2.0 & \textbf{63.4} & \textbf{61.3}  $\pm$0.5  & 0.528 & 0.519 & 0.971 & 0.921 \\
        4.0 & \textbf{63.0} & \textbf{60.9}  $\pm$0.5  & \textbf{0.533} & \textbf{0.524} & \textbf{0.989} & \textbf
        {0.938} \\
        8.0 & 62.1 & 60.1  $\pm$0.5  & 0.526 & 0.520 & 0.957 & 0.914 \\
       16.0 & 61.8 & 59.6  $\pm$0.5  & 0.530 & 0.519 & 0.965 & 0.912 \\
    \end{tabular}
  \end{center}
  \caption{The impact of $\lambda_{gen}$ on \mciccoco accuracy, together
with caption-generation performance (ROUGE-L and CIDEr against 5 references).
All results are obtained with a \seqff model (embedding size 2048 and
hidden-layer sizes of 2048 and 512).}
  \label{table:lambda_gen}
\end{table}

The results in Table~\ref{table:lambda_gen} illustrate one of the main points
of this paper.
That is, the ability to perform the comprehension task (as measured
by the accuracy metric) positively correlates with the ability to perform other tasks
that require machine comprehension, such as caption generation.
At $\lambda_{gen} = 4$, the \seqff model not only has a high accuracy of
detecting the ground-truth option, but it also generates its own captions given
the input image, with an accuracy measured on \mciccoco at 0.9890 (dev) and
0.9380 (test) CIDEr scores.
On the other hand, at an accuracy level of about 59\%
(on test, at $\lambda_{gen} = 0.1$), the generation performance is at only
0.9010 (dev) and 0.8650 (test) CIDEr scores.

We note that there is an inherent trade-off between prediction
accuracy and generation performance, as seen for $\lambda_{gen}$ values above 4.0.
This agrees with the intuition that training a \seqff model using
a loss $L(\thetab, \wb, \ub)$ with a larger $\lambda_{gen}$ means that
the ground-truth detection loss (the first term of the loss in Eq.\ref{eq:mloss})
may get overwhelmed by the word-generation loss (the second term).
However, our empirical results suggest that there is value in training models
with a multi-task setup, in which both the comprehension side as well as
the generation side are carefully tuned to maximize performance.


\section{Discussion}
We have proposed and described in detail a new multi-modal
machine comprehension task (DMC), combining the challenges of understanding
visual scenes and complex language constructs simultaneously.
The underlying hypothesis for this work is that
computer systems that can be shown to perform increasingly well
on this task will do so by constructing a visually-grounded
understanding of various linguistic elements and their dependencies.
This type of work can therefore benefit research in both machine visual understanding and
language comprehension.

The \seqff architecture that we propose for addressing this combined challenge
is a generic multi-task model.
It can be trained end-to-end to display both
the ability to choose the most likely text associated with an image
(thus enabling a direct measure of its ``comprehension'' performance),
as well as the ability to generate a complex description of that image
(thus enabling a direct measure of its performance in an end-to-end
complex and meaningful task).
The empirical results we present validate the underlying hypothesis of our work,
by showing that we can measure the decisions made by such a computer system
and validate that improvements in comprehension and generation happen in tandem.

The experiments presented in this work are done training our systems in an end-to-end
fashion, starting directly from raw pixels.
We hypothesize that our framework can be fruitfully used to show that
incorporating specialized vision systems (such as object detection, scene recognition,
pose detection, etc.) is beneficial.
More precisely, not only it can lead to a direct and measurable impact on a computer system's
ability to perform image understanding, but it can express that understanding in
an end-to-end complex task.



\end{document}